%% file: root.tex
\documentclass[letterpaper, 10 pt, conference]{ieeeconf}  

\IEEEoverridecommandlockouts                              

\usepackage{graphics} 
\usepackage{graphicx}
\usepackage{epsfig} 
\usepackage{mathptmx} 
\usepackage{times} 
\usepackage{amsmath} 
\usepackage{amssymb}  
\usepackage{caption}
\usepackage{subcaption}
\usepackage{bm}
\usepackage{booktabs,ragged2e}
\usepackage[flushleft]{threeparttable}
\usepackage{cuted}
\usepackage{cite}
\makeatletter
\let\NAT@parse\undefined
\makeatother
\usepackage{amsfonts}
\DeclareMathAlphabet{\mathcal}{OMS}{cmsy}{m}{n}
\usepackage{hyperref}
\hypersetup{
    colorlinks=true,       
    linkcolor=red,         
    citecolor=green,        
    filecolor=magenta,     
    urlcolor=cyan          
}
\usepackage{float}
\title{\LARGE \bf
Instance-Agnostic Geometry and Contact Dynamics Learning }

\author{Mengti Sun$^{*}$, Bowen Jiang$^{*}$, Bibit Bianchini, Camillo Jose Taylor, Michael Posa 
\thanks{*Authors have equal contribution.}
\thanks{All authors are affiliated with the GRASP Lab at the School of Engineering and Applied Science, University of Pennsylvania, Philadelphia, PA, 19104. \tt\small\{mengti, bwjiang, bibit, cjtaylor, posa\}@seas.upenn.edu}
}

\begin{document}

\maketitle
\thispagestyle{empty}
\pagestyle{empty}

\begin{strip}
    \vspace{-2.2cm}
    \centering
    \includegraphics[width=1.0\textwidth]{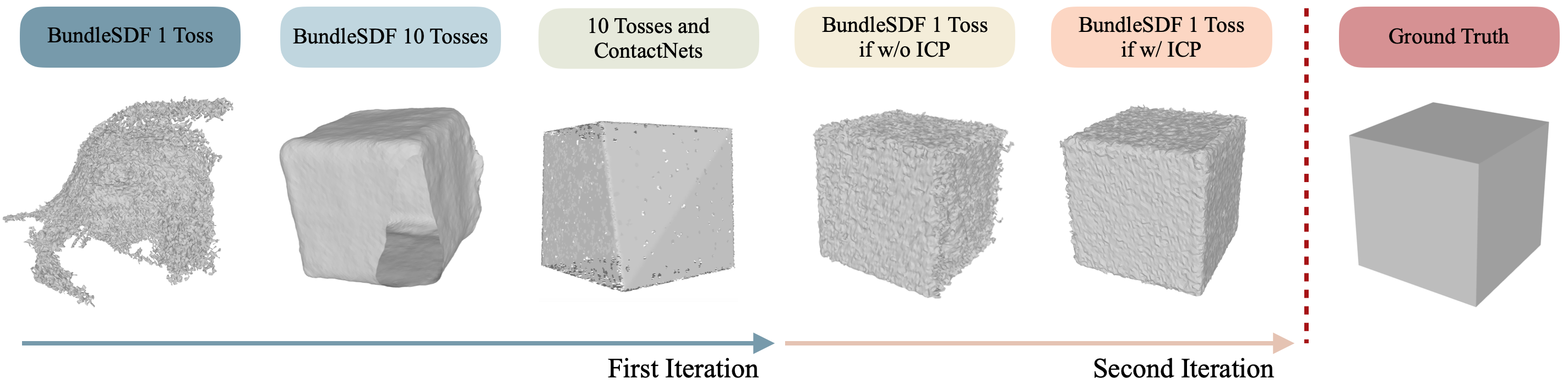}
    \captionof{figure}{Reconstructed geometries at different steps of the framework. The right-most geometry represents the final result, which significantly outperforms the initial reconstruction on one toss and even the reconstruction from ten tosses.}
    \label{fig:geo}
\end{strip}

\input{sections/abstract}
\input{sections/intro}
\input{sections/related}
\input{sections/approach}
\input{sections/exp}

\input{sections/discussion}


\addtolength{\textheight}{-12cm} 




\addtolength{\textheight}{12cm}
\clearpage
\bibliographystyle{IEEEtran}
\bibliography{IEEE}

\end{document}

%% file: sections/abstract.tex
\begin{abstract}
This work presents an instance-agnostic learning framework that fuses vision with dynamics to simultaneously learn shape, pose trajectories, and physical properties via the use of geometry as a shared representation. Unlike many contact learning approaches that assume motion capture input and a known shape prior for the collision model, our proposed framework learns an object's geometric and dynamic properties from RGBD video, without requiring either category-level or instance-level shape priors. We integrate a vision system, BundleSDF, with a dynamics system, ContactNets, and propose a cyclic training pipeline to use the output from the dynamics module to refine the poses and the geometry from the vision module, using perspective reprojection. Experiments demonstrate our framework's ability to learn the geometry and dynamics of rigid and convex objects and improve upon the current tracking framework.

\end{abstract}

%% file: sections/intro.tex
\section{INTRODUCTION}


Robots are increasingly being employed in complex real-world scenarios where they interact with a variety of objects. Vision systems are often employed to perform basic object detection and segmentation tasks ~\cite{girshick2015fast, carion2020end, hu2018learning, redmon2016you, zellers2018neural}. However successful manipulation often requires more detailed information about the object shape and relevant contact dynamics~\cite{pfrommer2021contactnets}.

This work explores an approach that combines an object modeling system with a system for refining geometry and modeling contact dynamics from data. It expands the horizon of robotic applications and allows us to contemplate systems that can manipulate previously unseen objects.


In the pursuit of estimating dynamics, recent research emphasizes the modeling of irregularities in an observed object trajectory, where the most interesting manipulation or frictional contact behaviors usually happen~\cite{pfrommer2021contactnets}. ContactNets~\cite{pfrommer2021contactnets}, as a state-of-the-art example, addresses this challenge from a motion capture perspective, obviating the need for force-sensing instruments. It optimizes inter-body distance functions and contact-frame Jacobians; however, it relies on having estimates for the object pose throughout the trajectory and a shape prior, such as a 3D mesh of the object, to learn a non-penetration loss function. Acquiring such a shape prior like a CAD model~\cite{labbe2020cosypose, wang2019normalized}, especially in open-world settings, can be labor intensive.


The BundleSDF~\cite{wen2023bundlesdf} framework is a state of the art system that can be used to build a 3D model of a moving object while simultaneously estimating its pose in space~\cite{liu2022gen6d, park2020latentfusion, wen2021bundletrack}.
This makes it an appropriate candidate to provide pose and shape priors to the ContactNets dynamics framework.


There are, however, some challenges associated with using BundleSDF in practice. When the trajectory is brief, pose estimation accuracy is high, but the distribution of viewpoints might be inadequate to achieve comprehensive geometry reconstruction. Besides, if the trajectory is lengthy, error accumulation compromises pose estimation and tracking accuracy, although the geometry reconstruction tends to be more precise. Motivated by this observation, our work is designed to learn a better shape representation, and then leverage it to improve pose estimation on shorter trajectories.


Our contribution can be summarized as follows:
\begin{enumerate}
    \item A novel integration method that use RGBD videos to produce a refined geometry and dynamics model from a physical learning framework.
    \item Design of a cyclic pipeline utilizing improved geometry to refine pose estimation on short trajectories, harnessing reprojection and iterative closest points.
    \item Experiments on a cube toss dataset collected with a manipulator that demonstrates improved pose estimation and geometry reconstruction over SOTA baselines. 
\end{enumerate}


\begin{figure*}[htbp] 
  \centering
  \includegraphics[width=\textwidth]{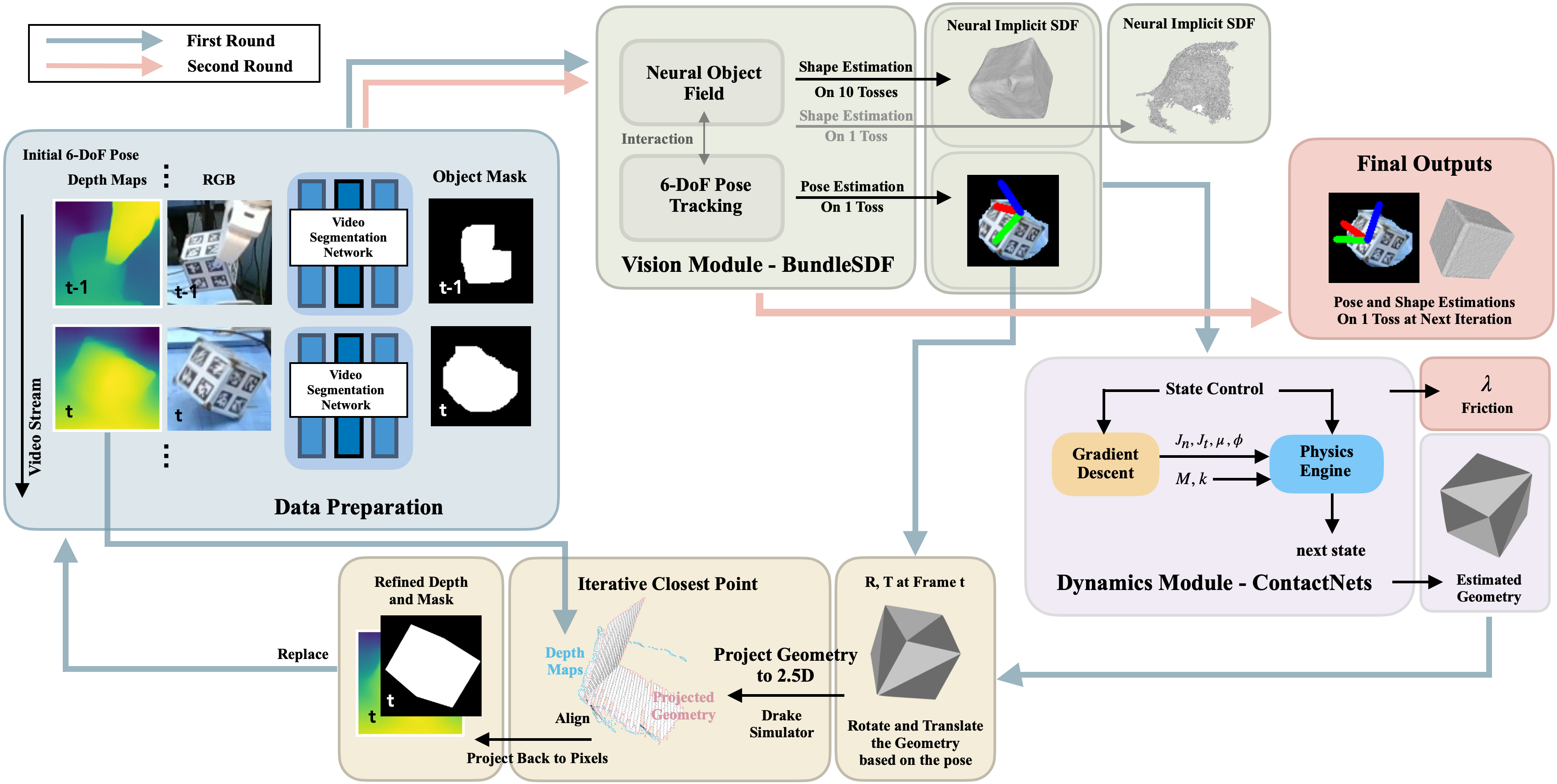}
  \caption{Framework overview. In the first round, BundleSDF predicts poses and geometry, and ContactNets refines the geometry and predicts frictional forces. The refined geometry is transformed based on the estimated pose and projected into the frame of the RGBD camera as a 2.5D point cloud. ICP is used to align this predicted depth and the raw depth in the dataset. The aligned geometry is projected to the pixel domain as a refined mask and depth. In the next round, BundleSDF can predict better poses and geometry from the refined dataset.}
  \label{fig:pipeline}
\end{figure*}


%% file: sections/related.tex
\section{RELATED WORK} \label{sec:rel}
\textbf{Neural Radiance Fields} NeRF~\cite{mildenhall2021nerf, xie2022neural} have shown impressive results in constructing  detailed 3D models of objects from a sparse set of views with zero 3D supervision. \cite{pumarola2021d, li2021neural, xian2021space, xian2021space} address the limitation of the original NeRF framework~\cite{mildenhall2021nerf} by introducing the temporal dimension to capture moving objects. 
BundleSDF~\cite{wen2023bundlesdf} utilizes an advanced NeRF model~\cite{mildenhall2021nerf} by combining geometry reconstruction with pose estimation, efficient ray sampling with depth-guided truncation for faster convergence, and normal-guided implicit regularization for smooth surface extraction.

\textbf{Contact Dynamics Learning} In the realm of Real-to-Simulation~\cite{shen2022acid, le2021differentiable, sundaresan2022diffcloud, jatavallabhula2021gradsim}, recent efforts focus on understanding and modeling the frictional contact dynamics of rigid bodies, estimating mass properties, inertia matrices, and the frictional forces in collisions. ContactNets~\cite{pfrommer2021contactnets} implicitly parameterizes the discontinuous contact behavior with a continuous inter-body signed distance function and contact-frame Jacobians. The system initializes the geometry from a shape prior, and optimizes it from trajectory measurements and a physics model. 
Relevant literature includes DANO~\cite{le2023differentiable} and GradSim~\cite{jatavallabhula2021gradsim}.
This manuscript considers a new approach by integrating BundleSDF~\cite{wen2023bundlesdf} to improve object modeling.

%% file: sections/approach.tex
\section{APPROACH} \label{sec:app}
We depict the overview of our approach in Figure~\ref{fig:pipeline}. The framework takes a collection of $n$ RGBD videos of the same object manipulated by a robotic arm, denoted as $\mathcal{V} = \left\{v_1, v_2, ..., v_n\right\}$. Each video sequence $v_{i} \in \mathcal{V}$ represents a toss of the rigid body colliding with a rigid flat surface.
The videos in $\mathcal{V}$ are consecutive in time, so they can be regarded as part of one longer video. Each frame in $v_{i} \in \mathcal{V}$ comprises an RGB image $\boldsymbol{I} \in \mathbb{R}^{3 \times r \times c}$ and a depth map $\boldsymbol{D} \in \mathbb{R}^{r \times c}$, where $r$ and $c$ denote the number of rows and columns. A binary mask $\boldsymbol{M} \in \mathbb{R}^{r \times c}$ and a starting pose of the object to be tracked in the camera frame are only required for the very first frame in the sequence, although optionally, binary masks for more frames can be provided for better instance segmentation.

\paragraph{Semi-Supervised Instance Segmentation}

Video object segmentation (VOS)~\cite{cheng2022xmem, oh2019video, duarte2019capsulevos} is an ideal approach to ensure the coherence of segmentation masks throughout a trajectory without labor-intensive annotations. Given the initial frame's segmentation mask of the object instance, we leverage the XMem~\cite{cheng2022xmem} model to automatically generate segmentation masks for the target object in all subsequent frames. It employs multiple distinct yet interconnected feature memory stores initialized from the initial frame. For each following frame, it retrieves information from these memories to generate a mask. This innovative approach yields high-quality features while conserving memory usage, and is particularly well-suited for the extended video sequences in our experiment.

\paragraph{Initial Estimation on a Single Video Clip}
This step runs BundleSDF on one video clip $v_{i}$ as a baseline. It finds visual feature correspondence between consecutive frames and collects a memory pool of keyframes that will participate in the pose graph optimization to ensure multi-view consistency and a Neural Object Field~\cite{wen2023bundlesdf} for geometry reconstruction. It generates initial pose estimates $\boldsymbol{p}$ for all frames in $v_{i}$ and a rough geometry reconstruction $\boldsymbol{O}_{0}$.

\paragraph{Geometry Reconstruction}
This step reconstructs a holistic geometry of the target object by sending the entire repertoire of video sequences in $\mathcal{V}$ to BundleSDF~\cite{wen2023bundlesdf}. Compared to an individual sequence $v_{i} \in \mathcal{V}$, a more extended sequence offers a wider range of viewpoints and contributes to more complete and smooth shape reconstructions. It also eliminates the occlusion problem in certain frames. 


The Neural Object Field represents shapes as level sets of a deep neural network $\Omega$ that learns geometry functions 
\begin{equation}
    \Omega: \boldsymbol{x} \in \mathbb{R}^{3} \mapsto s \in \mathbb{R}, \quad \left\{\boldsymbol{x} \in \mathbb{R}^3 \mid \Omega(\boldsymbol{x})=0\right\},
\end{equation}
where $\boldsymbol{x}$ is a 3D coordinate, $s$ is a signed distance value, and the zero set of the geometry function is the implicit surface of the object instance that can be derived from the first-order derivative of the neural network.

The optimization considers the uncertain free space, empty space, and near-surface space~\cite{wen2023bundlesdf} to handle imperfections in segmentation masks and depth maps. The Eikonal regularization $
L \varpropto \left(\|\nabla \Omega(\boldsymbol{x})\|_2-1\right)^2$~\cite{gropp2020implicit} is used to train the near-surface signed distance field. The framework outputs a geometry $\boldsymbol{O}$ describing the object instance in the 3D space.


\paragraph{Contact Dynamics Learning}
We now focus on a single video $v_{i} \in \mathcal{V}$. We let the ContactNets framework~\cite{bianchini2023simultaneously} take the 6-DoF pose estimations $\boldsymbol{p}_{0}$ in one video $v_{i}$ and reconstructed shape prior $\boldsymbol{O}$ from all videos as the inputs. 


ContactNets learns functional approximations of the inter-body distance $\Phi_{j}(\boldsymbol{q})$ with a geometric prior $\boldsymbol{O}$, the contact Jacobian $\boldsymbol{J}_{j}(\boldsymbol{q})$, and the friction coefficient $\mu_{j}$ using state transitions 
$\mathcal{D}=(\boldsymbol{x}_j, \boldsymbol{u}_j, \boldsymbol{x}_j^{\prime})_{j \in 1, \ldots, D}$, where $\boldsymbol{x}_j = [\boldsymbol{q}_j; \boldsymbol{v}_j]$ is the state of $j$-th contact frame, $\boldsymbol{q}_j$ contains the robot configuration and pose, $\boldsymbol{v}_j$ denotes the velocity, and $\boldsymbol{x}_j^{\prime}$ is the descrete-time dynamics of the system~\cite{bianchini2023simultaneously}. We follow the assumption in \cite{bianchini2023simultaneously} that the inertia quantities and non-contact impulses are known. The training loss function on the contact impulses $\boldsymbol{\lambda}_{j}$ and current state transition $\mathcal{L}(\boldsymbol{x}_j, \boldsymbol{u}_j, \boldsymbol{x}_j^{\prime}, \boldsymbol{\lambda}_{j})$ is established on prediction quality, contact activation, non-penetration criterion, and maximal dissipation~\cite{bianchini2023simultaneously}, and the problem is formulated as a tractable and convex program solved by gradient descent. 

The learned inter-body distance function $\Phi_{j}(\boldsymbol{q})$ outputs a 3D geometry $\tilde{\boldsymbol{O}}$ of the object instance, refined from the shape prior $\boldsymbol{O}$ with richer information from contact dynamics.

\paragraph{Cyclic Pipeline for Pose and Geometry Refinement}
We demonstrate a novel approach that interlaces the training of the BundleSDF and ContactNets frameworks in a cyclic fashion. Leveraging the geometry offered by ContactNets from full videos and contact dynamics, we devise a reprojection and iterative closest point (ICP) algorithm to allow BundleSDF to effectively refine its geometry and pose estimations in the next iteration. The framework can also be applied to a future video $v_{i^{\prime} > n}$ without relearning $\tilde{\boldsymbol{O}}$.


Utilizing the previous pose estimate $\boldsymbol{p}_{0}$ for each frame and the current point cloud $\tilde{\boldsymbol{O}}$, we can rotate and translate each point $\boldsymbol{w} \in \mathbb{R}^{3}$ in the 3D point cloud $\tilde{\boldsymbol{O}}$ to align it with the object instance in each frame of the video $v_i$, respectively.
\begin{equation} \label{eqn:trans}
	\tilde{\boldsymbol{O}}^{\prime} = \{  R \cdot \boldsymbol{w} + T \mid \forall \boldsymbol{w} \in \tilde{\boldsymbol{O}} \},
\end{equation}
where $R \in \mathbb{R}^{3 \times 3}$ and $T \in \mathbb{R}^{3}$ are the rotation and translation matrices obtained from $\boldsymbol{p}_{0}$, and $\cdot$ denotes matrix multiplication. We project the transformed 3D point cloud $\tilde{\boldsymbol{O}}^{\prime}$ to the image pixel domain using the Drake simulator \cite{drake} to generate a new set of depth maps and segmentation masks. 
\begin{align}
	& \boldsymbol{D}^{\prime} = \{ K \cdot \boldsymbol{w} \mid \forall \boldsymbol{w} \in \tilde{\boldsymbol{O}}^{\prime} \} \\
        & \boldsymbol{M}^{\prime} [u, v] =
            \begin{cases}
            1 & \text{if } \boldsymbol{D}^{\prime}[u, v] \ne \text{background} \\
            0 & \text{otherwise} \label{eqn:mask}
            \end{cases},
\end{align}
where $K \in \mathbb{R}^{3 \times 3}$ is the camera intrinsic matrix.

Here we rely on the initial pose estimates $\boldsymbol{p}_{0}$ derived from BundleSDF to transform the geometry $\tilde{\boldsymbol{O}}$, assuming that these poses are the pseudo ground-truth. Although the geometry $\tilde{\boldsymbol{O}}$ is better than the geometry from running BundleSDF on a single video clip, there could be a rotational and translational disparity between the modified depth maps $\boldsymbol{D}^{\prime}$ obtained from $\tilde{\boldsymbol{O}}^{\prime}$ and the original depth maps $\boldsymbol{D}$. Therefore, we implement an iterative closest point (ICP) algorithm to align these depth maps and bridge such disparities.

We project these two depth maps $\boldsymbol{D}$ and $\boldsymbol{D}^{\prime}$ into the RGBD camera view (also referred to as the 2.5D view to denote partial 3D information from depth maps) to create the source point cloud $\boldsymbol{O}_{s}$ and target point cloud $\boldsymbol{O}_{t}$. The ICP algorithm aims to find the corresponding points between two point clouds and a rotation $R$ and translation $T$ that minimizes their sum of squared distances.
\begin{equation}
    R, T =\arg \min _{R, T} \sum \left\|\boldsymbol{w}_{t}-R \cdot \boldsymbol{w}_{s} - T \right\|_2^2, \ \forall \boldsymbol{w}_{s} \in \boldsymbol{O}_{s}, \boldsymbol{w}_{t} \in \boldsymbol{O}_{t}
\end{equation}

The optimized $R$ and $T$ resulting from the ICP alignment process are employed to transform the source point cloud $\boldsymbol{O}_{s}$ in the 3D space. Subsequently, the transformed point cloud $\boldsymbol{O}^{\prime}_{s}$ is projected back to the pixel space, yielding refined depth maps $\boldsymbol{D}^{''}$ and segmentation masks $\boldsymbol{M}^{''}$, similar to the steps outlined in Equations~\ref{eqn:trans}-\ref{eqn:mask}. At this moment, the refined set of depth maps $\boldsymbol{D}^{''}$,  segmentation masks $\boldsymbol{M}^{''}$, as well as the original RGB images and starting pose from the initial frame constitute a new dataset. This newly formed dataset is then subjected to another iteration of the BundleSDF framework. This iterative procedure results in further enhanced pose estimates and geometry reconstructions.

%% file: sections/exp.tex
\section{EXPERIMENTS} \label{sec:exp}

\begin{figure}[htp] 
  \centering
  \includegraphics[width=0.45\textwidth]{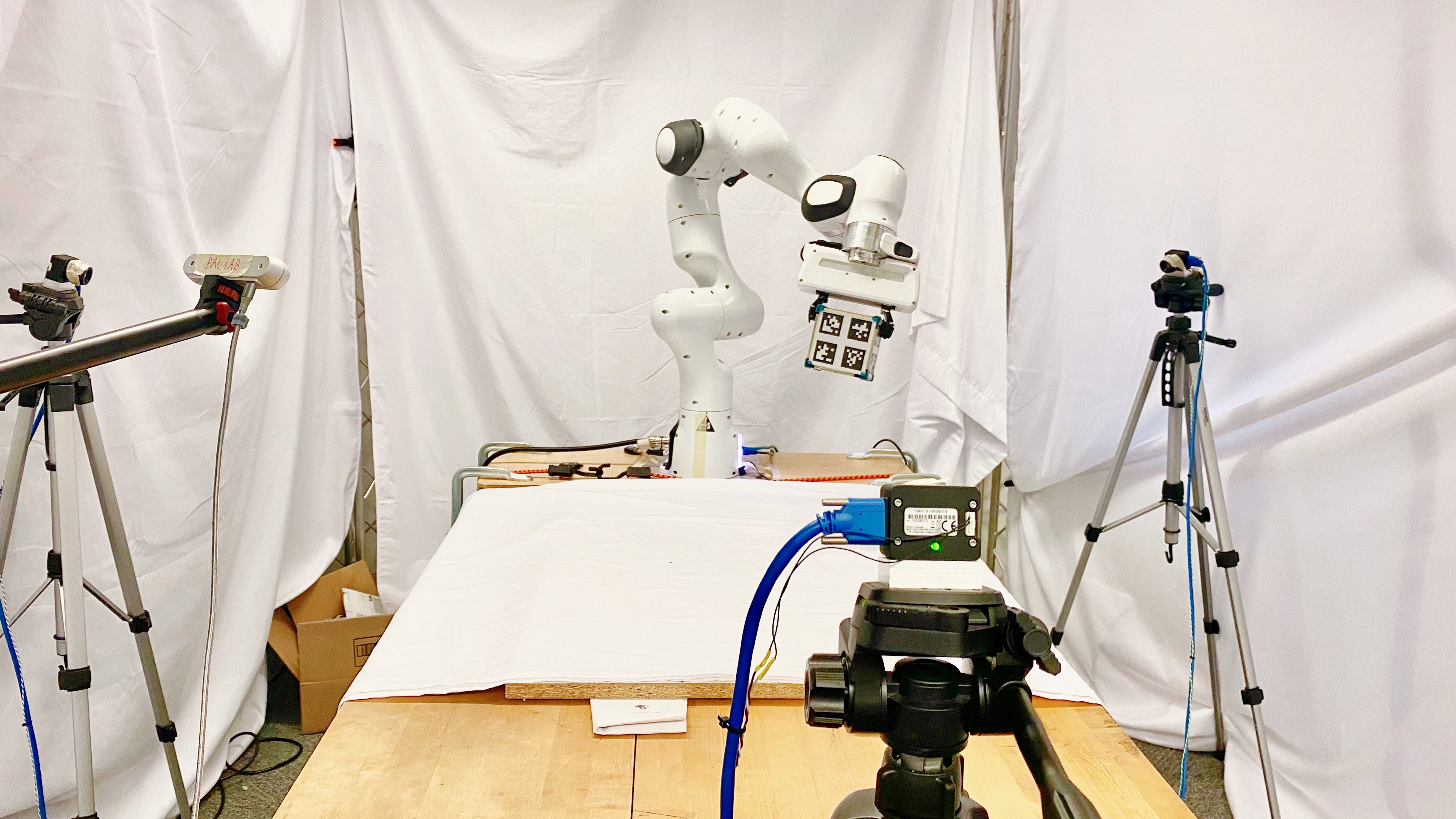}
  \caption{Experimental setup used in the data collection process showcasing the robotic arm grasping the cube for tossing.}
  \label{fig:setup}
\end{figure}

We test our framework on a robot cube toss dataset we collected: a rigid cube affixed with AprilTags is tossed by a Franka Panda 7-DoF robot arm onto a flat rigid surface while data is captured by a 30Hz IntelRealSense D455 RGBD camera. The cube is tracked by three 30Hz PointGrey cameras using TagSLAM~\cite{DBLP:journals/corr/abs-1910-00679} to establish ground-truth poses.

Numerical results and ablation studies are shown in Table~\ref{tab:results}. We evaluate pose estimation and geometry reconstruction separately as in ~\cite{wen2023bundlesdf}: For 6-DoF object pose, we compute 1) area under the curve (AUC) percentage of \textbf{ADD}. 2) AUC percentage of \textbf{ADD-S}.
\begin{align}
    &\textit{ADD} = \frac{1}{|\mathcal{M}|}\sum\limits_{\boldsymbol{w}\in \mathcal{M}} \| (R\cdot\boldsymbol{w}+T)- (\tilde{R}\cdot \boldsymbol{w}+\tilde{T})\|_2 \\
    &\hspace{-2mm}\textit{ADD-S} = \frac{1}{|\mathcal{M}|} \sum\limits_{\boldsymbol{w}_1 \in \mathcal{M}} \min\limits_{\boldsymbol{w}_2 \in \mathcal{M}} \| (R\cdot\boldsymbol{w}_1+T)- (\tilde{R}\cdot\boldsymbol{w}_2+\tilde{T})\|_2 \hspace{-1mm}
\end{align} 
where the tilde denotes ground-truth $R$ and $T$ and $\mathcal{M}$ being object's ground-truth geometry. 3) success rate \textbf{SR} under $5^\circ5$cm metric~\cite{wen2023bundlesdf}, namely orientation error within $5^\circ$ and translation error within $5$ cm. For geometry reconstruction, we compute the Chamfer distance \textbf{CD} = \vspace{1mm} $\frac{1}{2|\mathcal{M}_1|}\sum\limits_{\boldsymbol{w}_1 \in \mathcal{M}_1} \min\limits_{\boldsymbol{w}_2 \in \mathcal{M}_2} \|\boldsymbol{w}_1 - \boldsymbol{w}_2\|_2 + \frac{1}{2|\mathcal{M}_2|}\sum\limits_{\boldsymbol{w}_2 \in \mathcal{M}_2}  \min\limits_{\boldsymbol{w}_1\in \mathcal{M}_1} \|\boldsymbol{w}_1  - \boldsymbol{w}_2\|_2$ between the generated and ground-truth meshes.\vspace{0.5mm}

The last row of Table~\ref{tab:results} indicates the results achieved by our framework, which exhibits a stronger performance than the baselines. We posit that the depth maps $\boldsymbol{D}^{''}$ and masks $\boldsymbol{M}^{''}$ derived from the refined geometry exhibit the capability to effectively recover occlusions and defective masks better than the original depth $\boldsymbol{D}$ and masks $\boldsymbol{M}$. Besides, the final reconstructed geometry, as shown in Figure~\ref{fig:geo}, significantly outperforms the original BundleSDF, regardless of whether it is running on 1 toss or 10 tosses. It's intriguing that the entire system has no prior knowledge of the geometry, yet it still recovers the cube almost flawlessly.

In the ablation study of the cyclic training pipeline, having the cycle will decrease the CD score to 0.18cm with better pose estimation simultaneously. Besides, compared to BundleSDF on 10 tosses, the reconstructed geometry undergoes a refinement process by ContactNets with dynamic contact learning, which reduces CD from 1.75cm to 1.18cm. 

When the ContactNets module is omitted, we directly reproject the geometry from running BundleSDF on 10 tosses to get $\boldsymbol{M}^{''}$ and $\boldsymbol{D}^{''}$, bypassing ContactNets. It shows that the process of learning contact dynamics exerts a positive influence on both pose estimations and geometry reconstruction by BundleSDF in the next iteration.



\begin{figure}[htb]
    \centering
    \begin{subfigure}[b]{0.5\textwidth}
        \centering
        \includegraphics[width=0.475\linewidth]{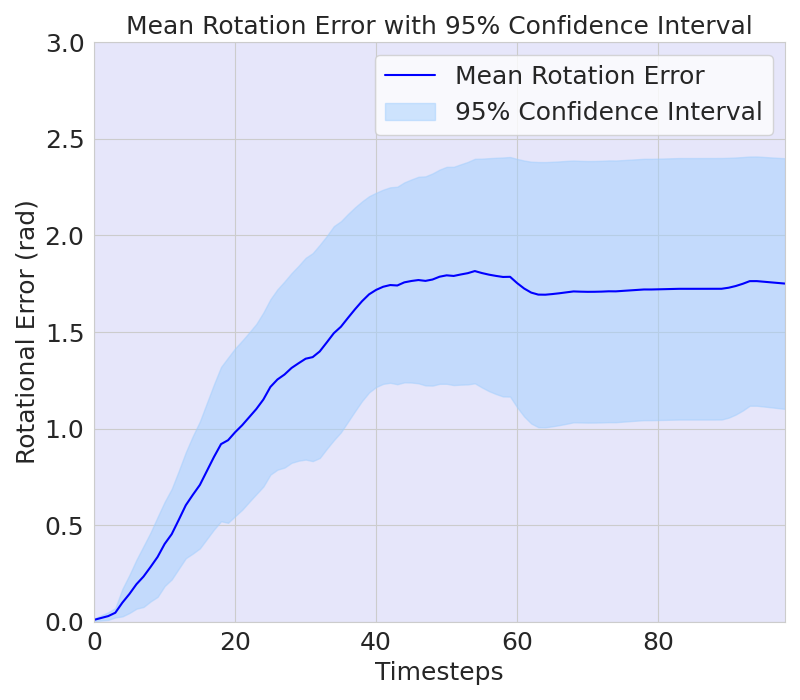}%
        \hfill
        \includegraphics[width=0.475\linewidth]{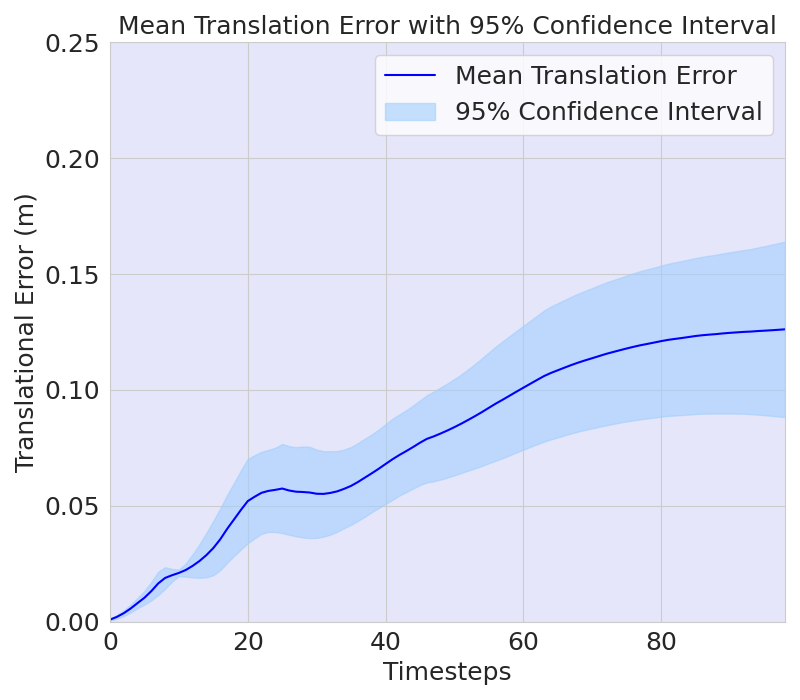}
    \end{subfigure}
    \caption{Estimated rollouts from input initial condition and ContactNets model trained with a dataset of 9 tosses. The rotation and translation errors are compared with the input trajectory from BundleSDF output assuming as ground truth.}
    \label{fig:cnets}
\end{figure}

We also evaluate the ContactNets output by comparing the estimated rollouts
with the ground-truth trajectories. The training dataset comprises a total of 9 trajectories, which are pose estimations from BundleSDF. As illustrated in Figure~\ref{fig:cnets}, we present the mean rotational error and translational error along with a 95\% confidence interval across 10 trajectories and rollouts of 99 steps. This underscores ContactNets' ability to generate accurate trajectory predictions within the constraints of a relatively modest dataset.

\begin{table}[tp]
\begin{threeparttable}
\caption{Experiment Results and Ablation Studies}
\label{tab:results}
\setlength\tabcolsep{0pt} 

\begin{tabular*}{\columnwidth}{@{\extracolsep{\fill}} l cccc}
\toprule
      & 
     \multicolumn{3}{c}{Pose} & \multicolumn{1}{c}{Reconstruction} \\ 
\cmidrule(l{1em}r{1em}){2-4}\cmidrule(l{4em}){4-5}
     & ADD-S(\%) $\uparrow$ & ADD(\%) $\uparrow$  & SR(\%) $\uparrow$ &  CD(cm) $\downarrow$ \tnote{[a]}\\
\midrule
     BundleSDF 1 toss \tnote{[b]} & 65.02 & 58.08 & 40.17 & 34.20 \\ 
     BundleSDF 10 tosses \tnote{[c]} & 64.89 &32.42 & 32.10 & 1.75 \\
     Ours w/o cyclic pipeline \tnote{[d]} & 65.02 & 58.08 & 40.17 & 1.18 \\
     Ours w/o ContactNets \tnote{[e]} & 52.60 & 48.36 & 41.83 & 0.96\\
     Ours w/o ICP \tnote{[f]} & 65.43 & 60.45 & 48.20 &0.74 \\
     Ours \tnote{[g]} & \textbf{69.50} & \textbf{68.40} & \textbf{49.86} & \textbf{0.18} \\
\addlinespace
\bottomrule
\end{tabular*}
\begin{tablenotes}
\item[a] $\uparrow$ means we prefer higher scores. $\downarrow$ means we prefer lower scores.
\item[b] Run BundleSDF on one video clip of 1 toss;
\item[c] Run BundleSDF on the full video of 10 tosses;
\item[d] Run ContactNets using poses from BundleSDF on 1 toss and geometry from BundleSDF on 10 tosses. Compared to the final result, it has no cyclic pipeline. Compared to the first baseline, it has an identical trajectory;
\item[e] Run BundleSDF on 10 tosses to get geometry, refine geometry via ICP, and run BundleSDF again on 1 toss using this geometry in a cyclic fashion. Compared to the final result, geometry is not refined by ContactNets;
\item[f] Run ContactNets using poses from BundleSDF on 1 toss and geometry on 10 tosses, and run BundleSDF again on 1 toss using geometry from ContactNets in a cyclic fashion. Compared to the final result, poses are not refined by ICP;
\item[g] Run ContactNets using poses from BundleSDF on 1 toss and geometry from BundleSDF on 10 tosses, align geometry from ContactNets via ICP, and run BundleSDF again on 1 toss using aligned geometry. Final result.
\vspace{-3mm}

\end{tablenotes}
\end{threeparttable}
\end{table}

%% file: sections/discussion.tex
\section{DISCUSSION AND FUTURE WORK} \label{sec:dis}
Our study introduces a robust framework that performs instance-agnostic geometry and contact dynamics learning. The cornerstone of our approach lies in integrating two crucial components: the BundleSDF~\cite{wen2023bundlesdf} pose estimation and geometry reconstruction framework, in conjunction with the ContactNets~\cite{pfrommer2021contactnets} contact learning framework.
The approach resolves the lack of motion capture data and shape priors for contact dynamics learning. 
Experiments with ablation studies demonstrate the effectiveness of our system. 
Moving forward, we are extending our preliminary efforts toward continuing the cyclic pipeline to convergence, unifying loss functions for joint optimizations, and encompassing a wider variety of novel and nonconvex objects in the open world.
